\documentclass{article}

\usepackage[final]{nips_2017}

\usepackage{times}
\usepackage{graphicx}
\usepackage{subfigure}

\usepackage{natbib}

\usepackage{algorithm}
\usepackage{algorithmic}

\usepackage[usenames,dvipsnames]{color}
\usepackage{url}
\usepackage{amsmath}
\usepackage{amsthm}
\usepackage{amssymb}
\usepackage{amsfonts,eucal,amsbsy}
\usepackage{mathtools}
\usepackage{multirow}
\usepackage{caption}

\newcommand{\ignore}[1]{}
\newcommand{\vect}[1]{\mathbf{#1}}
\newcommand{\struct}[1]{\boldsymbol{#1}}

\newcommand{\argmax}{\mathrm{argmax}}

\begin{document}

\title{Generative and Discriminative Text Classification \\with Recurrent Neural Networks}

\author{
  Dani Yogatama, Chris Dyer, Wang Ling, and Phil Blunsom \\
  DeepMind\\
  \texttt{\{dyogatama,cdyer,lingwang,pblunsom\}@google.com}
}

\maketitle

\begin{abstract}
We empirically characterize the performance of discriminative and generative LSTM models for text classification. We find that although RNN-based generative models are more powerful than their bag-of-words ancestors (e.g., they account for conditional dependencies across words in a document), they have higher asymptotic error rates than discriminatively trained RNN models. However we also find that generative models approach their asymptotic error rate more rapidly than their discriminative counterparts---the same pattern that \citet{ngandjordan} proved holds for linear classification models that make more na\"{\i}ve conditional independence assumptions.  Building on this finding, we hypothesize that RNN-based generative classification models will be more robust to shifts in the data distribution. This hypothesis is confirmed in a series of experiments in zero-shot and continual learning settings that show that generative models substantially outperform discriminative models.
\end{abstract}

\section{Introduction}
Neural network models used in natural language processing 
applications are usually trained discriminatively.
This strategy succeeds for many applications when training data is abundant and the data distribution is stable. Unfortunately, neural networks require a lot of training data, and they tend to generalize poorly when the data distribution shifts (e.g., new labels, new domains, new tasks).
\ignore{However, humans can solve many language understanding problems with very limited numbers
of training examples, and cope robustly with data distribution shifts.
For example, we need not see tens of thousands of spam emails
to be able to classify a new email as spam or not spam.
While our ability to use knowledge acquired  
from other tasks plays a
major role in our generalization ability, it is also
clear that we can learn from fewer examples even in
a task where we have no prior knowledge.}
In this paper, we explore using generative models to obtain improvements in sample complexity and ability to adapt to shifting data distributions.

While neural networks are traditionally used as discriminative models~\citep{ney95,rubinstein:1997}, their flexibility makes them well suited to estimating class priors and class-conditional observation likelihoods. We focus on a simple NLP task---text classification---using discriminative and generative variant models based on a common neural network architecture (\S\ref{sec:models}). These models use an LSTM~\citep{lstm} to process documents as sequences of words. In the generative model, documents are generated word by word, conditioned on a learned class embedding; in the discriminative model the LSTM ``reads'' the document and uses its hidden representation to model the class posterior. In contrast to previous generative models for text classification, ours can model unbounded (conditional) dependencies among words in each document.

We demonstrate empirically that our discriminative model obtains a lower asymptotic error rate than its generative counterpart, but it approaches this rate more slowly~(\S\ref{sec:complexity}). This behavior is precisely the pattern that \citet{ngandjordan}
proved will hold in general for generative and discriminative \emph{linear} models. Finding the same pattern with our models is somewhat surprising since our generative models are substantially more powerful than the linear models analyzed in that work (e.g., they model conditional dependencies among input features), and because their theoretical analysis relied heavily on linearity.

Encouraged by this result, we turn to learning problems where good sample complexity is crucial for success and explore whether generative models might be preferable to discriminative ones.
We first consider the single-task continual learning setting 
in which the labels (classes) are introduced sequentially,
and we can only learn from the newly introduced examples~(\S\ref{sec:continual}).
Discriminative models are
known to suffer from catastrophic forgetting when learning
sequentially from examples from a single class at a time,
and specialized techniques are actively being developed 
to minimize this problem \citep{prognets, elastic, pathnet}.
Generative models, on the other hand, are a more 
natural fit for this kind of setup since
the maximization of the training 
objective for a new class can be decoupled
from other classes more easily (e.g., parameters of a na\"{\i}ve Bayes classifier can be estimated
independently for each class).
In order to compare discriminative and generative models
more fairly,
we use a generative model that shares many parameters across classes and evaluate its performance in this
setting. 

\begin{figure*}[t]
    \vspace{-0.7cm}
    \centering
    \subfigure[Discriminative]{\includegraphics[scale=0.65]{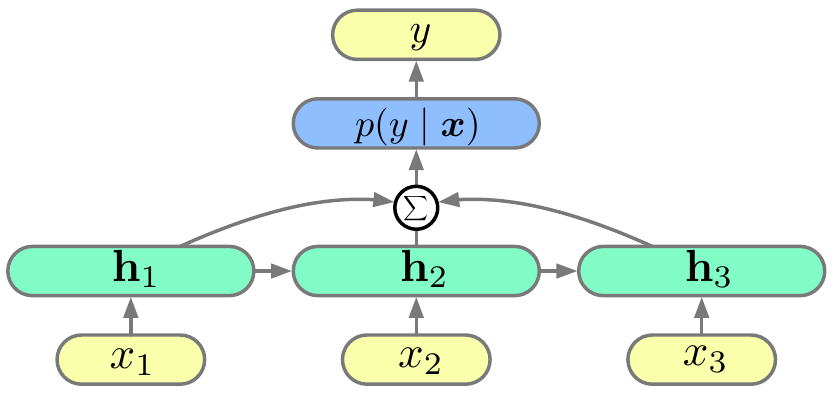}}
    \hspace{0.5cm}
    \subfigure[Generative]{\includegraphics[scale=0.65]{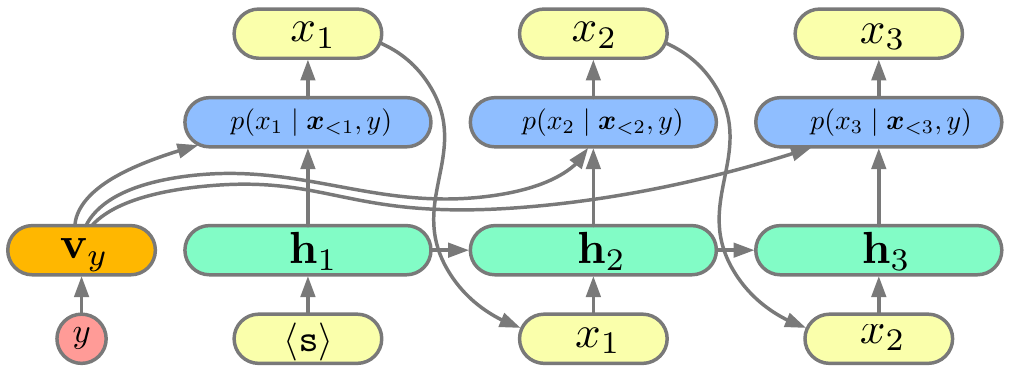}}
    \vspace{-0.3cm}
    \caption{Illustrations of our discriminative (left) and generative (right) LSTM models.}
    \label{fig:models}
    \vspace{-0.5cm}
\end{figure*}

Finally, we compare the performance of 
discriminative and generative LSTM language models for
zero-shot learning, where we construct 
a semantic label space that
is fixed during training based on an auxiliary task (\S{\ref{sec:zeroshot}}).
We investigate whether learning to map documents
onto this semantic space (discriminative training) or
learning to generate from points in the semantic space 
(generative training) is better. Here, we find substantial benefits for generative models.

\vspace{-0.3cm}
\section{Models}
\vspace{-0.3cm}
\label{sec:models}
Inputs to a text classification system are a document $\struct{x} = \{x_1, x_2, \ldots, x_T\}$, where $T$ is its length in words, and it will predict a label $y \in \mathcal{Y}$.
We compare discriminative and generative text classification models. Discriminative models are trained to distinguish 
the correct label among possible choices.
Given a collection of 
labeled documents $\{(\struct{x}_i, y_i)\}_{i=1}^N$, these models are trained to maximize the conditional probability
of the labels given the documents: $\sum_{i=1}^N \log p(y_i\mid\struct{x}_i)$.
Generative models, on the other hand, are trained
to maximize the joint probability of labels and documents under the following factorization:
$\sum_{i=1}^N \log p(\struct{x}_i, y_i) = \sum_{i=1}^N \log p(\struct{x}_i\mid y_i)p(y_i)$. When predictions are made, Bayes' rule is used to compute $p(y \mid \boldsymbol{x})$.

In both models, we represent a word $x$ by a $D$-dimensional embedding  $\vect{x} \in \mathbb{R}^D$. Figure~\ref{fig:models} shows an illustration of our models, and we describe them in details in the following section.

\subsection{Discriminative Model}
Our discriminative model uses LSTM with ``peephole'' connections to encode a document and
build a classifier on top of the encoder by using the average of the
LSTM hidden representations as the document representation.

Specifically, given an input word embedding $\vect{x}_t$,
we compute its hidden representation $\vect{h}_t \in \mathbb{R}^E$ with LSTM as follows:
\begin{align*}
\vect{i}_t &= \sigma(\vect{W}_{i} [\vect{x}_t; \vect{h}_{t-1};\vect{c}_{t-1}] + \vect{b}_i) &\vect{f}_t &= \sigma(\vect{W}_{f} [\vect{x}_t; \vect{h}_{t-1}; \vect{c}_{t-1}] + \vect{b}_f)\\
\vect{c}_t &= \vect{f}_t \odot \vect{c}_{t-1} + \vect{i}_t \odot \text{tanh}(\vect{W}_{c} [\vect{x}_t ; \vect{h}_{t-1}] + \vect{b}_c) &\vect{o}_t &= \sigma(\vect{W}_{o} [\vect{x}_t;  \vect{h}_{t-1} ; \vect{c}_{t}] + \vect{b}_o)\\
\vect{h}_t &= \vect{o}_t \odot \text{tanh}(\vect{c}_t),
\end{align*}
where $[\mathbf{u};\mathbf{v}]$ denotes vector concatenation.
We then add a softmax layer on top of this LSTM, so
the probability of predicting a label $y \in \mathcal{Y}$ is: $p(y \mid \struct{x}) \propto \exp( (\frac{1}{T} \sum_{t=0}^T \vect{h}_t^{\top}) \vect{v}_y + b_y)$,
where $\vect{V}\in\mathbb{R}^{E \times |\mathcal{Y}|}$ is the softmax parameters 
and $\vect{b} \in \mathbb{R}^{|\mathcal{Y}|}$ is the bias.
We use a simple average of LSTM hidden representations 
since in our preliminary experiments it works better
than using the last hidden state $\vect{h}_T$, 
and it is computationally much cheaper for long documents (hundreds of words) 
than attention-based models.
Importantly, this model is trained discriminatively to maximize 
the conditional probability of the label given the document: $p(y\mid\struct{x};\vect{W},\vect{V})$.

\subsection{Generative Models}
Our generative model is a class-based language model, shown in Figure~\ref{fig:models}.
Here, we similarly compute hidden representation $\vect{h}_t$ with LSTM.
Additionally, we also have a label 
embedding matrix $\vect{V}\in \mathbb{R}^{E\times|\mathcal{Y}|}$. We use the chain rule to factorize the probability $p(\boldsymbol{x} \mid y)$ into a sequential prediction:
$p(\boldsymbol{x} \mid y) = \prod_{t=1}^{T} p(x_t \mid \boldsymbol{x}_{<t}, y)$.
To predict the word $x_{t}$,
we concatenate the LSTM's hidden representation  $\boldsymbol{x}_{<t}$ (called $\vect{h}_t$) with
the label embedding $\vect{v}_{y}$ and add a softmax layer over 
vocabulary with parameters $\vect{U}$ and class-specific bias parameters $\vect{b}_y$: $p(x_{t} \mid \boldsymbol{x}_{<t}, y) \propto \exp(\vect{u}_{x_{t}}^{\top} [\vect{h}_{t}; \vect{v}_y] + b_{y, x_{t}})$.
We designate this model ``Shared LSTM'', since it shares some parameters across classes (i.e., 
the word embedding matrix, LSTM parameters $\vect{W}$, and softmax parameters $\vect{U}$).
This model's novelty
owes to the fact that there is a single conditional model that shares parameters
whose behavior is modulated by the given label embedding, whereas in traditional generative classification
models, each label has an independent LM associated with it,
such as the generative $n$-gram language classification models in \citet{pengandschuurmans2003}.

In addition to the above model, we also experiment with a class-based generative language model
where there is no shared component among classes (i.e., every class has its own word embedding, LSTM,
and softmax parameters). 
One benefit of this approach is that training can be parallelized across 
classes, although the resulting model has larger number of parameters.
We denote this model by ``Independent LSTMs''.

Note that the underlying LSTM of both our generative models
is similar to the discriminative model,
except that it is trained to maximize the 
joint probability $p(y, \struct{x}; \vect{W},\vect{V},\vect{U}) = p(\struct{x}\mid{y}; \vect{W},\vect{V},\vect{U}) p(y)$.
In terms of the number of parameters, these generative models have extra parameters $\vect{U}$
that are needed to predict words (we can view the label embedding matrix as a substitute for the
softmax parameter in the discriminative case).
For prediction, we compute 
$\hat{y} = \argmax_{y \in \mathcal{Y}} p(\struct{x}\mid{y}; \vect{W},\vect{V},\vect{U}) p(y)$ using the empirical relative frequency estimate of $p(y)$.

\vspace{-0.2cm}
\section{Experiments}
\label{sec:experiments}
\vspace{-0.2cm}
\subsection{Datasets}
We use publicly available datasets from \citet{zhangcnn} to evaluate our models (\footnotesize\url{http://goo.gl/JyCnZq}\normalsize). 
They are standard text classification datasets that include news classification, sentiment analysis, Wikipedia article classification, and questions and answers categorization.
Table~\ref{tbl:datasets} shows descriptive statistics of datasets used in our experiments.
For each dataset, we randomly hold 5,000 examples from the original training set to be used as our development set.


\begin{table*}[h]
\vspace{-0.2cm}
\caption{
Descriptive statistics of datasets used in our experiments.
\label{tbl:datasets}}
\vspace{-0.2cm}
\centering
\footnotesize
\begin{tabular}{|c||r|r|r|r|}
\hline
Name & \#Train & \#Dev & \# Test & \# Classes \\
\hline
AGNews & 115,000 & 5,000 & 7,600 & 4 \\
Sogou & 445,000 & 5,000 & 60,000 & 5 \\
Yelp & 645,000 & 5,000 & 50,000 & 5 \\
Yelp Binary & 555,000 & 5,000 & 7,600 & 2 \\
DBPedia & 555,000 & 5,000 & 70,000 & 14 \\
Yahoo & 1,395,000 & 5,000 & 60,000 & 10 \\
\hline
\end{tabular}
\vspace{-0.7cm}
\end{table*}

\subsection{Baselines}
In addition to our generative vs. discriminative models in \S{\ref{sec:models}} and baselines from previous work on these datasets,
we also compare with the following generative models:
\paragraph{Na\"{\i}ve Bayes classifier.}
A simple count-based unigram language
model that uses na\"{\i}ve Bayes assumption to factorize $p(\struct{x}\mid y) = \prod_{t=1}^T p(x_t\mid y)$.
\paragraph{Kneser--Ney Bayes classifier.}
A more sophisticated count-based language model
that uses trigrams and Kneser--Ney smoothing $p(\struct{x}\mid y) = \prod_{t=1}^T p(x_t \mid x_{t-1}, x_{t-2}, y)$.
Similar to the na\"{\i}ve Bayes classifier, we construct one language model per class and predict by computing:
$\hat{y} = \argmax_{y \in \mathcal{Y}} p(\struct{x}\mid y) p(y)$.
\paragraph{Na\"{\i}ve Bayes neural network.}
Last, we also design a na\"{\i}ve Bayes baseline where $p(x_t \mid y)$ is modeled by a feedforward neural network
(in our case, we use a two layer neural network).
This is an extension of the na\"{\i}ve Bayes baseline, where we replace the class-conditional count-based unigram language model
with a class-conditional vector-based unigram language model.

\subsection{Implementation Details}
In all our experiments, we set the word embedding dimension $D$ and the LSTM hidden dimension $E$ to 100.\footnote{We also experimented with
setting both $D$ and $E$ to 50 and 300, they resulted in comparable performance.}
For the generative model, the dimension of the class embedding is also set to 100.
We train our model using AdaGrad \citep{adagrad} and tune the learning rate on development sets.
We also use the development sets to decide when to stop training based on classification accuracy as the evaluation metric.

\subsection{Sample Complexity and Asymptotic Errors}
\label{sec:complexity}
\citet{ngandjordan} theoretically and empirically show that generative linear models reach their (higher) asymptotic error faster
than discriminative models (na\"{\i}ve Bayes classifier vs. logistic regression).
While it is difficult to derive the theoretical properties of expressive recurrent neural network
models such as ours, we empirically evaluate the performance of these models.

Table~\ref{tbl:allresults} summarizes 
our results  on the full datasets, along with results from previous work on these datasets.
Our discriminative LSTM model is competitive with 
other discriminative models based on logistic regression \citep{zhangcnn, fasttext}
or convolutional neural networks \citep{zhangcnn, charcrnn, vdcnn}.
All of the generative models have lower classification accuracies.
These results agree with \citet{ngandjordan} 
that discriminative models have lower asymptotic errors than generative models.

Comparing various generative models,
we can see that the generative LSTM models
are generally better than baseline generative
models with stronger independence assumptions (i.e., na\"{\i}ve Bayes, Kneser--Ney Bayes, and na\"{\i}ve Bayes neural network).
Our results suggest that LSTM is an effective method to 
capture dependencies among words in a document.
We also compare the two 
generative LSTM models: shared LSTM and independent LSTMs.
The results are roughly similar.

\begin{figure*}[t]
    \centering
    \subfigure{\includegraphics[scale=0.24]{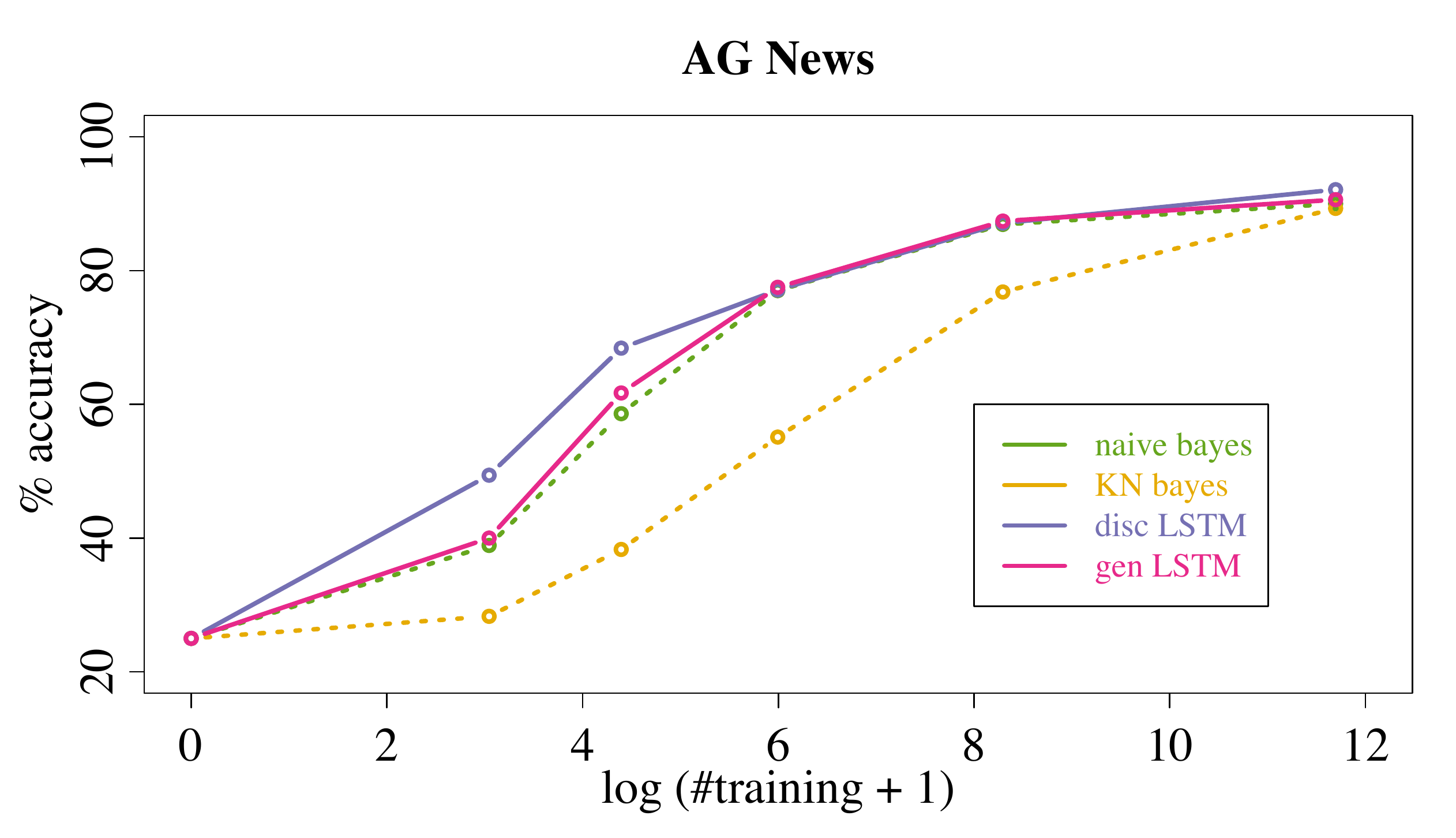}}
    \subfigure{\includegraphics[scale=0.24]{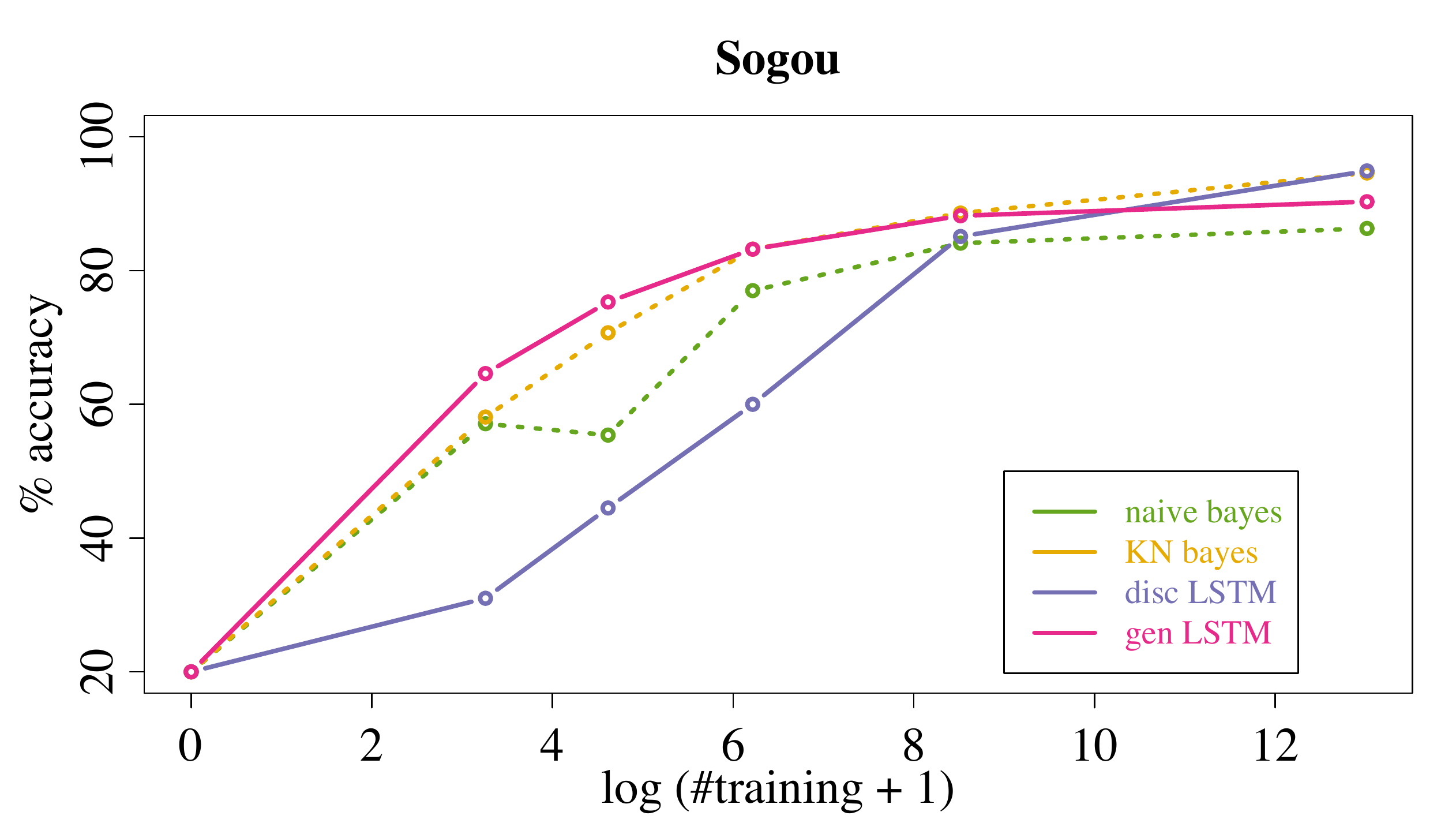}}
    \subfigure{\includegraphics[scale=0.24]{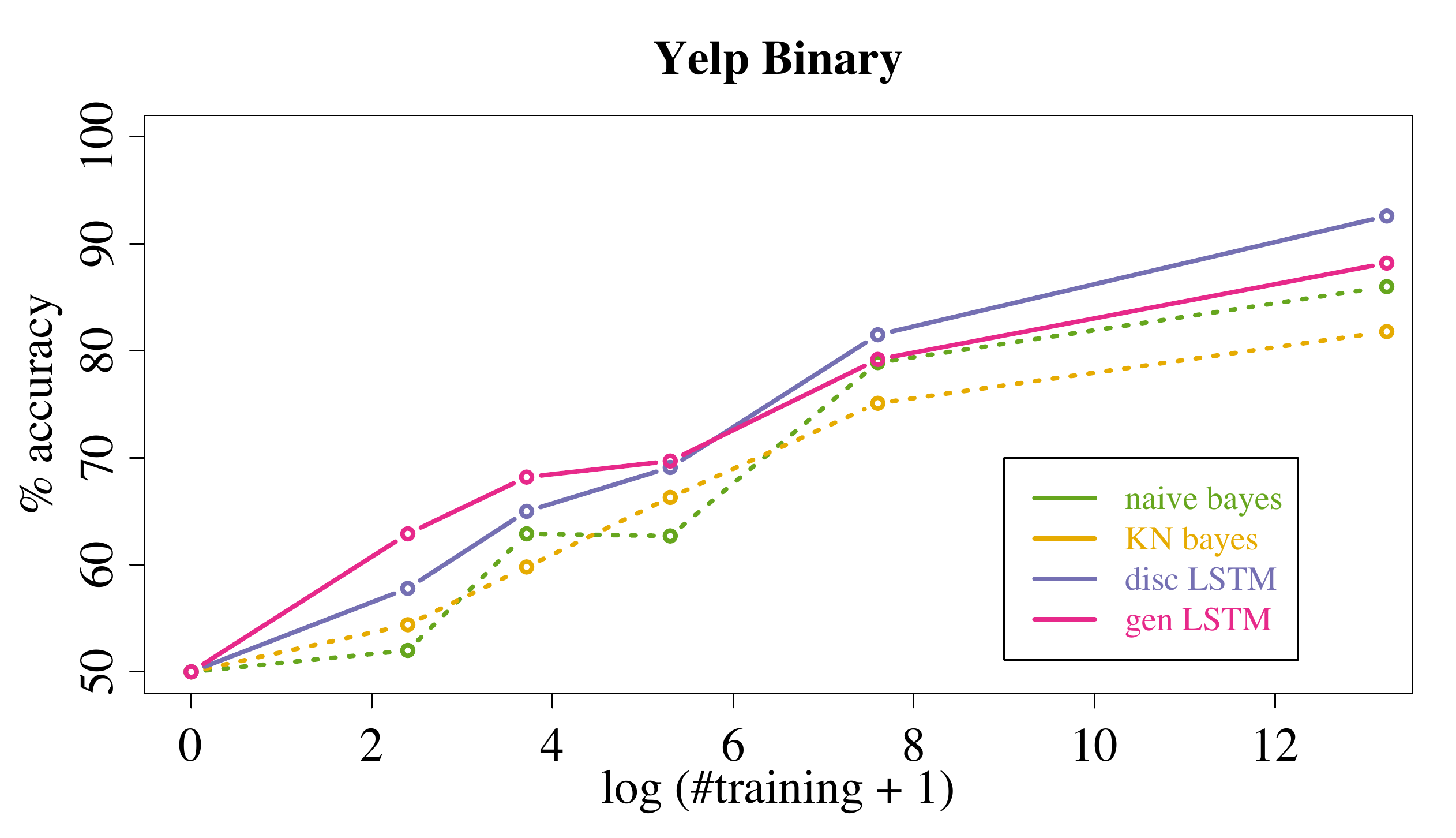}}
    \subfigure{\includegraphics[scale=0.24]{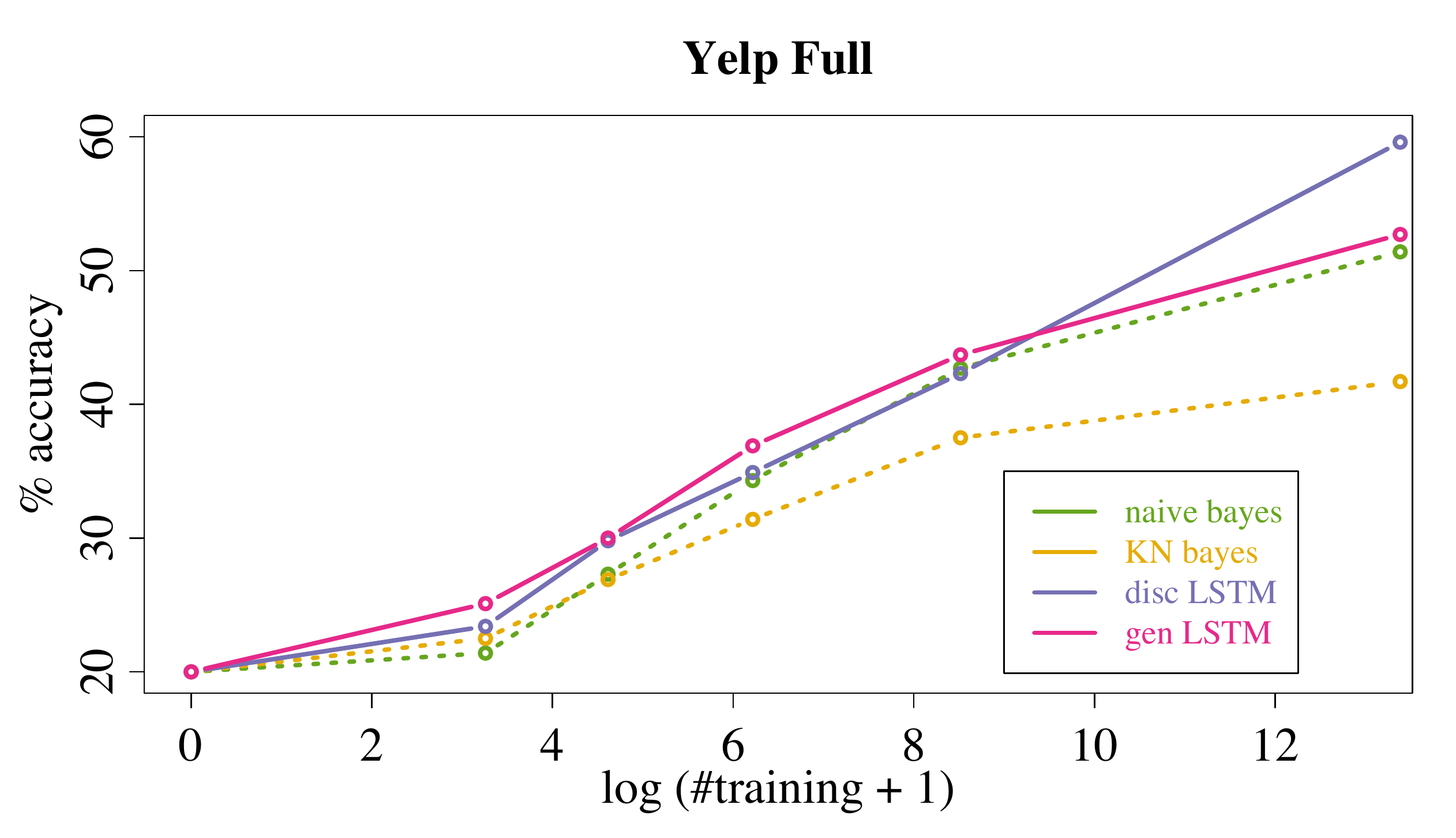}}
    \subfigure{\includegraphics[scale=0.24]{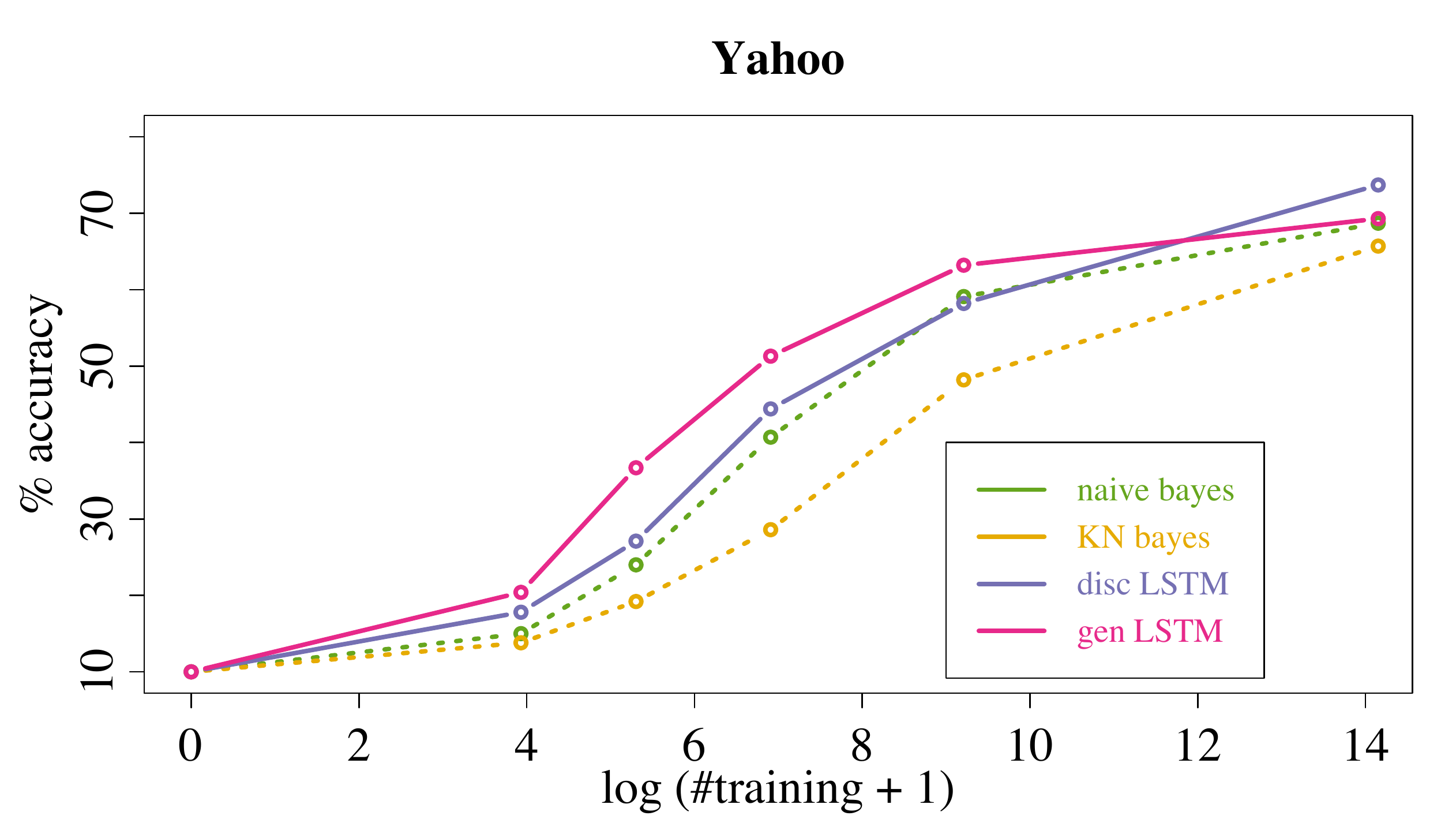}}
    \subfigure{\includegraphics[scale=0.24]{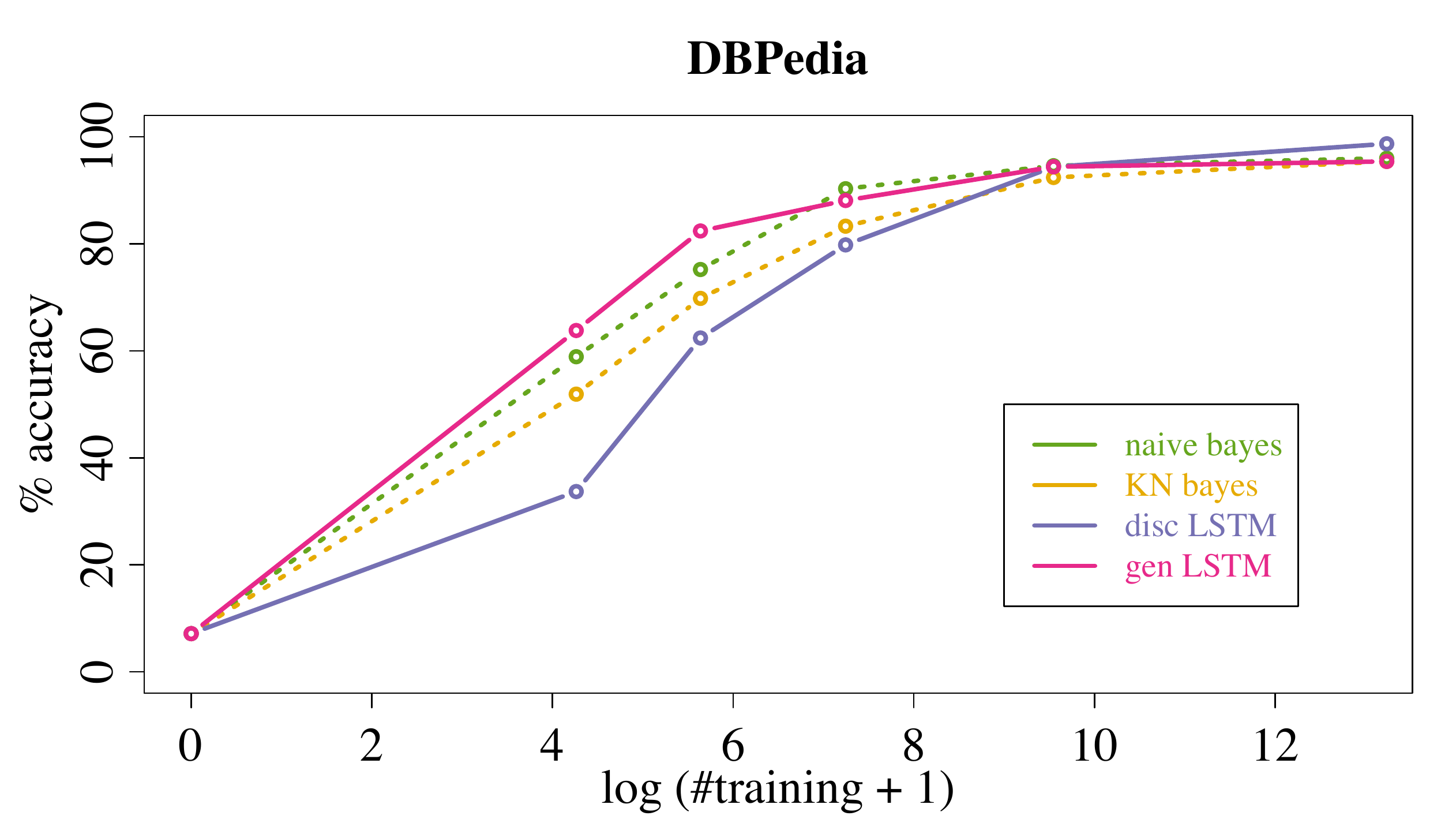}}
\vspace{-0.5cm}
    \caption{Accuracies of generative and discriminative models with varying training size.}
    \label{fig:plots}
\vspace{-0.7cm}
\end{figure*}

Next, we evaluate our models with varying training size.
For each of our six datasets, we randomly choose 5, 20, 100, and 1000 examples \emph{per class}.
We train the models on 
these smaller datasets and report results in Figure~\ref{fig:plots}.
Our results show that the generative shared LSTM model outperforms the discriminative model in
almost all cases in the small-data regime on all datasets
except one (AG News).
Among generative models, the 
generative LSTM model still achieves better 
classification accuracies compared to 
na\"{\i}ve Bayes and Kneser--Ney Bayes models, even in the
small-data regime.
While it is difficult to analyze the theoretical 
sample complexity of deep recurrent models, we
see this collection of results 
as an empirical support that
generative nonlinear models 
have lower sample complexity 
than their discriminative counterparts.

\begin{table*}[t]
\caption{
Summary of results on the full datasets.
\label{tbl:allresults}
}
\vspace{-0.1cm}
\centering \footnotesize
\begin{tabular}{|c||r|r|r|r|r|r|r||r|r|}
\hline
{Models} & {AGNews} & {Sogou} & {Yelp Bin} & {Yelp Full} &{DBPed} & {Yahoo} \\
\hline
\hline
Na\"{\i}ve Bayes & 90.0 & 86.3 & 86.0 & 51.4 & 96.0 & 68.7 \\
Kneser--Ney Bayes & 89.3& 94.6 & 81.8 & 41.7 & 95.4 & 69.3 \\
MLP Na\"{\i}ve Bayes & 89.9 & 76.1 & 73.6 & 40.4 & 87.2 & 60.6 \\ 
Discriminative LSTM & 92.1 & 94.9 & 92.6 & 59.6 & 98.7 & 73.7 \\
Generative LSTM--independent comp. & 90.7 & 93.5 & 90.0 & 51.9 & 94.8 & 70.5 \\
Generative LSTM--shared comp. & 90.6 & 90.3 & 88.2 & 52.7 & 95.4 & 69.3 \\
\hline
bag of words \citep{zhangcnn} & 88.8 & 92.9 & 92.2 & 58.0 & 96.6 & 68.9 \\
fastText \citep{fasttext} & 92.5 & 96.8 & 95.7 & 63.9 & 98.6 & 72.3 \\
char-CNN \citep{zhangcnn} & 87.2 & 95.1 & 94.7 & 62.0 & 98.3 & 71.2\\
char-CRNN \citep{charcrnn} & 91.4 & 95.2 & 94.5 & 61.8 & 98.6 & 71.7\\
very deep CNN \citep{vdcnn} &  91.3 & 96.8 & 95.7 & 64.7 & 98.7 & 73.4 \\ 
\hline
\end{tabular}
\vspace{-0.5cm}
\end{table*}

\subsection{Continual learning}
\label{sec:continual}
Our next set of experiments investigate properties of
discriminative and generative LSTM models 
to adapt to data distribution shifts.
An example of data distribution shift is when
new classes are introduced to the models.
In the real-world setting, being able to detect
emergence of a new class and 
train on examples
from a new class that shows up at a later time 
without having to retrain the model
on the entire dataset is extremely attractive, especially
in cases where we have a very large dataset and model.
We focus on how well these models learn from 
new classes in this (sub)section
and discuss detection of data distribution shifts in 
more details in \S{\ref{sec:discussion}}.

\paragraph{Setup.} We consider the setup where models are
presented with examples from each class sequentially (single-task continual learning).\footnote{Here, we focus on a single-task continual learning setup, although the idea can be
generalized to a multitask setting as well.}
Here, each model has to learn information
from newly introduced examples to be able to correctly
classify documents into this new class,
but they cannot train from previously seen classes
while doing so.
Table~\ref{tbl:contresults} summarizes our
results and we discuss them in details in the followings.

\paragraph{Discriminative LSTM.} We first investigate the performance of the 
LSTM discriminative model.
Discriminative models are known to suffer from catastrophic forgetting---where the models overtrain 
on the newly introduced class and fail to retain
useful information from previous classes---in this setup.
In these experiments, every time 
we see examples from a new class,
we update all parameters of the model 
with information the new examples.
However, even after extensive tuning of learning 
rate and freezing some components of the network,\footnote{For example, we experiment with fixing the word embedding matrix and train all other components and fixing the LSTM language
model component after seeing the first class and only train the
softmax parameters.} 
we were unable to avoid catastrophic forgetting.
We observe that since the model is trained to \emph{discriminate} among possible classes, when it only sees examples
from a single class for tens or hundreds of iterations, 
it adjusts its parameters to always predict the new class.
Of course, in theory, there might be an oracle learning
rate that would prevent catastrophic forgetting while still
acquiring enough knowledge about the newly introduced classes
to update the model.
However, we find that in practice it is very difficult 
to discover these learning rate values, especially for a reasonably large LSTM model that takes high-dimensional input such as a long news article.
Even for the same learning rate value, the development set performance varies widely across multiple training runs.
Note that since this model is trained discriminatively,
it is also not trivial make use of information from unlabeled data
to pretrain some components of the model, except for the word embedding matrix.
A promising method to prevent catastrophic forgetting in discriminative models is elastic weight consolidation \citep{elastic}. However, the method requires computing a Fisher information matrix, and it is not clear how to compute it efficiently for complex models such as LSTM on GPUs.

\begin{table}[t]
\vspace{-0.1cm}
\begin{minipage}[t]{0.48\linewidth}
\centering \small
\vspace{-2.8cm}
\begin{tabular}{|c|r|r|r|}
\hline
{Datasets} & {Shared-Gen} & {Ind.-Gen} & {Disc.} \\
\hline
\hline
AG News & 90.2 & 90.7 & 40.5 \\
Yelp Full & 51.4 & 52.7 & 20.0 \\
Yelp Binary & 86.4 & 90.0 & 57.2 \\
DBPedia & 95.7 & 94.8 & 8.3 \\
Yahoo & 68.5 & 70.5 & 10.0 \\
\hline
\end{tabular}
\vspace{0.55cm}
\caption{Continual learning results. Shared and Ind.-Gen are the generative shared and independent models respectively (see text for details). \label{tbl:contresults}}
\end{minipage}
\hfill
\begin{minipage}[t]{0.46\linewidth}
\centering
\includegraphics[scale=0.22]{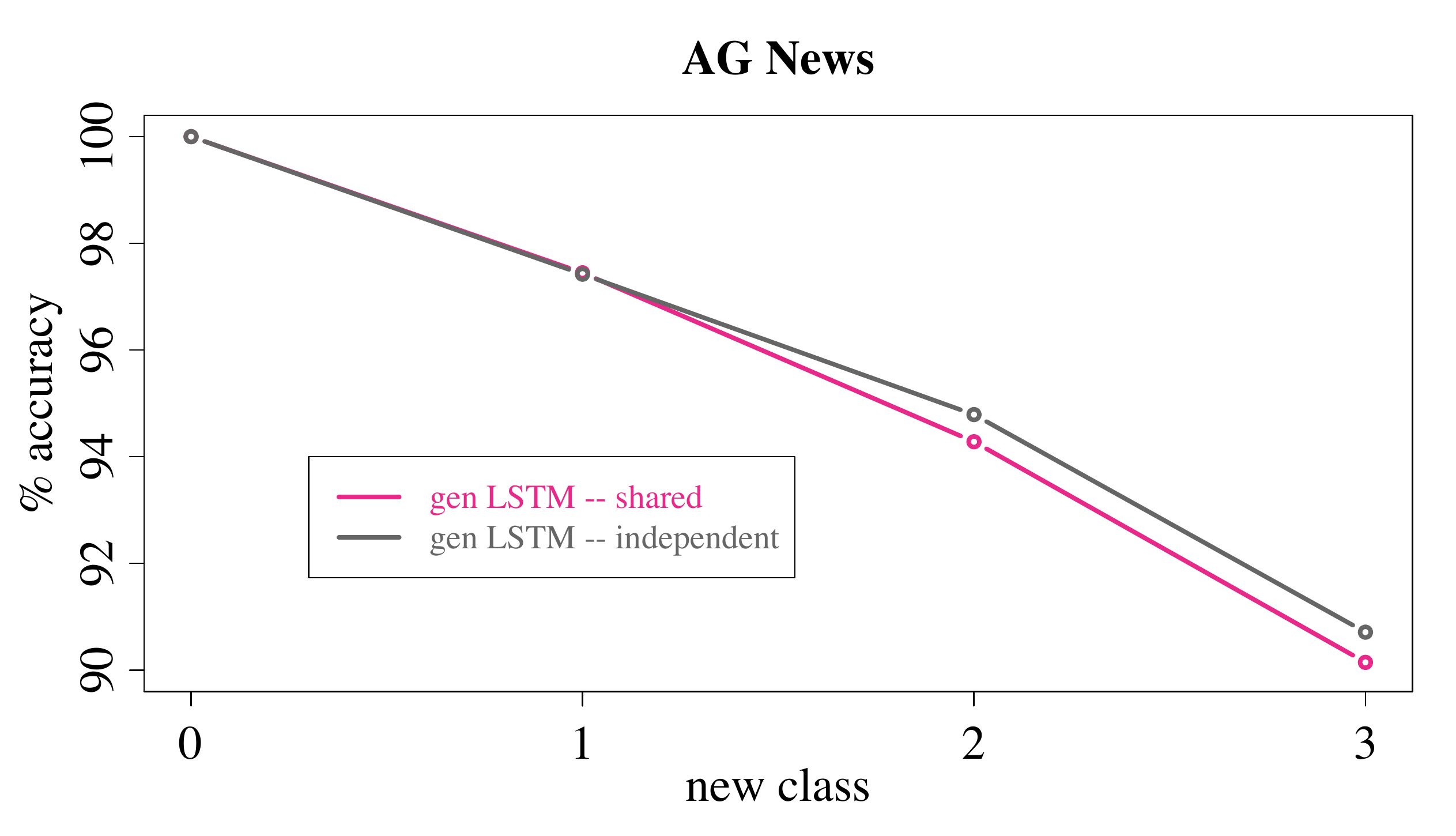}
\captionof{figure}{Classification accuracies on the AG News dataset for the generative LSTM models as we introduce class 0, 1, 2, and 3.
\label{fig:continualplot}}
\end{minipage}
\vspace{-1.2cm}
\end{table}

\vspace{-0.3cm}
\paragraph{Generative LSTM.} 
Next, we consider the generative independent LSTMs model.
Parameter estimations of 
some generative models, such as the na\"{\i}ve Bayes 
and this independent LSTMs, can be naturally decoupled across
classes. 
As a result, these models can easily incorporate information from
newly introduced examples from a new class.
Every time a new class is introduced, 
we simply learn a new model of $p(\boldsymbol{x}\mid y_{\text{new}}; \vect{W}_{y_{\text{new}}}, \vect{v}_{y_{\text{new}}}, \vect{U}_{y_{\text{new}}})$
and $\forall y \in  \mathcal{Y}$, update $p(y)$.
One possible drawback of this approach is that the size of
the model grows with the number of classes.

Last, we experiment with the generative shared LSTM model.
Our training procedure for this model is as follows. 
We first train the LSTM language model 
part on a large amount of unlabeled data.\footnote{In practice, we train on the 
corresponding training dataset 
without using document labels.}
When training the language model on unlabeled data, 
we remove the class-specific bias component $\vect{b}_y$ 
and set the class embedding $\vect{v}_y$
to be a random vector $\vect{\tilde{y}}$ with a bounded norm $\Vert \vect{\tilde{v}}_y \Vert_2 \leq 1$. 
After we pretrain the shared components, 
we freeze their parameters 
and tune the class embedding $\vect{v}_y$ as well 
as the class-specific softmax bias $\vect{b}_y$ on 
the labeled training data.
The benefit of this training---compared 
to having a separate LSTM model for each class---is that 
it is faster to train (in the presence of examples from a new class).
Given a new class $y$, we only need to 
learn two vectors: $\vect{v}_y$ and $\vect{b}_y$.
We can see from results in Table~\ref{tbl:contresults} and 
Figure~\ref{fig:continualplot} that the generative
shared LSTM model trained with this procedure approaches 
the performance of its equivalent model 
that can see examples from all classes, and performs competitively with the generative independent LSTMs model.

\vspace{-0.2cm}
\subsection{Zero-shot learning}
\vspace{-0.2cm}
\label{sec:zeroshot}
Our last set of experiments compare the performance of 
discriminative and generative LSTM language models for
zero-shot learning, where the label embedding space
is fixed based on an auxiliary task. 
Humans can acquire new concepts and learn relations
among these concepts from an external task, and
use this knowledge effectively across multiple tasks.
In these experiments, we use datasets where the class
labels are semantically meaningful concepts (e.g., \texttt{science}, \texttt{sports}, \texttt{business}---instead of star ratings from 1 to 5).

\vspace{-0.2cm}
\paragraph{Setup.}
We remove labels of documents
from one of the classes, but we provide the models
with knowledge about the classes
from external sources in the form of class embeddings $\vect{V}$.
For example, when labels are words
(e.g. \texttt{science}, \texttt{sports}, \texttt{business}, etc.), we construct a semantic space
by learning the label embeddings $\vect{v}_y$ 
using standard word embedding techniques for all $y \in \mathcal{Y}$.
In order to do this, we use pretrained GloVe word
embedding vectors \citep{glove}.\footnote{\url{ http://nlp.stanford.edu/projects/glove/}}
In cases where the class labels 
consist of more than one words
(e.g., society and culture), we choose one 
word from the labels (e.g., society).
At test time, we see examples from all classes
and evaluate precision and recall of the hidden class,
as well as the overall accuracy on all classes.

\vspace{-0.2cm}
\paragraph{Discriminative LSTM.}
The model learns from the labeled data to place documents in the semantic space, such that embeddings of documents are close to embeddings
of their respective labels.
In practice, we fix the softmax 
parameters $\vect{V}$ and learn an embedding of the
document to maximize $\exp( (\frac{1}{T} \sum_{t=0}^T \vect{h}_t^{\top}) \vect{v}_y)$, where $\vect{h}_t$ is the LSTM hidden state for word $t$ in the document.
In our experiments, this model never predicts
the hidden class (zero precision and recall).
Our results show that while discriminative 
training of an expressive model such as LSTM
on high dimensional text data produces
a reliable predictor of seen classes,
the resulting model overfits to the seen classes.

\vspace{-0.4cm}
\paragraph{Generative LSTM.}
The model learns to 
generate from points in the label semantic space 
using the labeled documents.
The model may infer how 
to generate a document about \texttt{politics} 
without ever having 
seen such an example in the training data.
Similar to the discriminative case, we fix
$\vect{V}$, which in this case plays the role of
class embeddings, and train other parts of the model
on all training data except examples from the hidden class (we use the generative shared LSTM since naturally the generative independent
LSTMs cannot be used in this experiment).
We observe on the development set that this model
is able to predict examples from the unseen class
with high precision, but very low recall ($\approx 1\%$).
We design a self-training algorithm that add 
predicted hidden class examples 
from the development set
to the training set and allow the model to train
on these predicted examples.
We show the results in Table~\ref{tbl:zeroshotallresults}.
We can see that for most hidden classes, 
the generative model achieves good 
performance.
For example, on the AG News dataset, the model
performs reasonably well for any of the hidden classes.
For a more difficult dataset 
where the overall accuracy is not very high such as Yahoo, the precision of the hidden class is lower, 
and as a result the recall also suffers. 
Nonetheless, the model is still able to
achieve reasonable overall accuracy in some cases (recall that the accuracy 
of this model trained on the full dataset 
without any hidden class is 69.3).

Of course,
if we include predicted hidden class examples 
to the training set of a discriminative model, it can also achieve good performance on all classes.
However, the main point is that the discriminative LSTM
model \emph{never} predicts the hidden classes
without any training data.

\begin{table}[t]
\vspace{-0.5cm}
\caption{
Zero shot learning results on four datasets.
Hidden class indicates the class that is not included
in the training data. We show precision and recall on test data for the hidden class, as well as accuracy
for examples from all classes.
\label{tbl:zeroshotallresults}
}
\centering \footnotesize
\begin{tabular}{|c||l||r|r|r|}
\hline
{Dataset} & {Hidden Class} & {Prec.} & {Recall} & {Acc.} \\
\hline
\hline
 \multirow{4}{*}{\bf AG News} & world & 94.4 & 77.8 & 87.8 \\
& sports  & 95.7 & 83.3 & 85.5 \\
& business & 84.9 & 60.1 &83.9 \\
& science and tech & 92.0 & 54.3& 83.2\\
\hline
\multirow{5}{*}{\bf Sogou} & sports & 95.0 & 80.5 & 87.4 \\
& finance & 24.2 & 0.7 & 73.1 \\
& entertainment & 90.2 & 78.8 & 86.6 \\
& automobile & 42.9 & 0.7 & 72.1 \\
& science and tech & 99.7 & 58.7 & 85.6 \\
\hline
\multirow{10}{*}{\bf Yahoo} &society and culture & 42.8 & 7.9 & 64.9 \\
&science and math & 48.3 & 9.8 & 63.2 \\
&health & 26.3 & 0.4 & 61.8 \\
&education and reference & 23.5 & 3.8 & 65.2 \\
&computers and internet & 45.4 & 3 & 60.8 \\
&sports & 52.9 & 52.9 & 64.6 \\
&business and finance & 43.6 & 17.3 & 66.2 \\
&entertainment and music & 44.9 & 2.3 & 63.2 \\
&family and relationships &8.3& 0.05& 62.5\\
&politics and government & 48.6& 10.4 & 62.1 \\
\hline
\multirow{14}{*}{\bf DBPedia} & company & 98.9 & 46.6 & 93.3 \\
& educational institution & 99.2 & 49.5 & 92.8 \\
& artist & 88.3 & 4.3 & 90.3\\
& athlete & 96.5& 90.1 & 94.6\\
& office holder & 0 & 0 & 89.1 \\
& mean of transportation & 96.5 & 74.3 & 94.2\\
& building & 99.9 & 37.7 & 92.1 \\
& natural place & 98.9 & 88.2& 95.4 \\
& village & 99.9& 68.1 & 93.8 \\
& animal & 99.7& 68.1 & 93.8 \\
& plant & 99.2 & 76.9 & 94.3 \\
& album & 0.03 & 0.001 & 88.8 \\
& film & 99.4 & 73.3 & 94.5 \\
& written work & 93.8 & 26.5 & 91.3\\
\hline
\end{tabular}
\vspace{-0.5cm}
\end{table}

\vspace{-0.4cm}
\paragraph{Two-class zero-shot learning.}
We also perform experiments with the generative LSTM model 
when we hide two classes
on the AG News dataset and show the results in Table~\ref{tbl:zeroshotallresultstwoclass}.
In this case, the model does not perform as well
since the precision of predicting hidden classes
drops significantly, introducing too much noise
in the training data. However, the model is still
able to learn some useful information since the
overall accuracy is still higher than 50\% (the accuracy of models that are only trained on two classes without any zero-shot learning of the hidden classes).

\begin{table}[t]
\vspace{-0.3cm}
\caption{
Zero-shot learning results with two hidden class on the AG News dataset. We show P0 (P1) and R0 (R1) 
that indicate precision and recall for hidden class one (two), as well as overall accuracy.
\label{tbl:zeroshotallresultstwoclass}
}
\vspace{-0.3cm}
\centering \footnotesize
\begin{tabular}{|c||r|r|r|r|r|}
\hline
{Classes} & {P0} & {R0} & {P1} & {R1} & {Acc.} \\
\hline
\hline
world+sports & 43.2 & 3.4 & 54.7 & 90.2 & 67.2 \\
world+business & 55.4 & 75.6 & 25.9 & 2.2 & 67.6 \\
world+science/tech & 40.5 & 5.7 & 38.7 & 47.8 & 61.1 \\
sports+business & 62.3 & 80.7 & 48.3 & 6.6 & 67.6\\
sports+science/tech & 66.2 & 85.5 & 66.8 & 6.7 & 67.6 \\
business+science/tech & 43.6& 62.1& 59.0& 1.9 & 63.3 \\
\hline
\end{tabular}
\vspace{-0.5cm}
\end{table}

\vspace{-0.3cm}
\section{Discussion}
\label{sec:discussion}
\vspace{-0.2cm}
\paragraph{Computational complexity.}
In terms of training and inference time, discriminative models are much faster. For example, for our smallest (biggest)
dataset that contains 115,000 (1,395,000) training
examples, it takes approximately two hours (two days) to get good generative
models, whereas training the discriminative models only takes
approximately 20 minutes (6 hours).
The main drawback of generative models in NLP applications
is in the softmax computation, since it has to be
done over an entire vocabulary set.
In many cases, such as in our experiments,
the size of the vocabulary is in the order of hundreds
of thousands. There are approximate methods to speed up 
this softmax computation, such as via hierarchical 
softmax \citep{hiersoftmax}, noise contrastive estimation \citep{nce}, sampled softmax \citep{sampledsoftmax}, or one-vs-each approximation \citep{oneeach}. 
However, even with these approximations,
discriminative models are still much faster.

\begin{figure}[h]
\vspace{-0.4cm}
    \centering
    \includegraphics[scale=0.23]{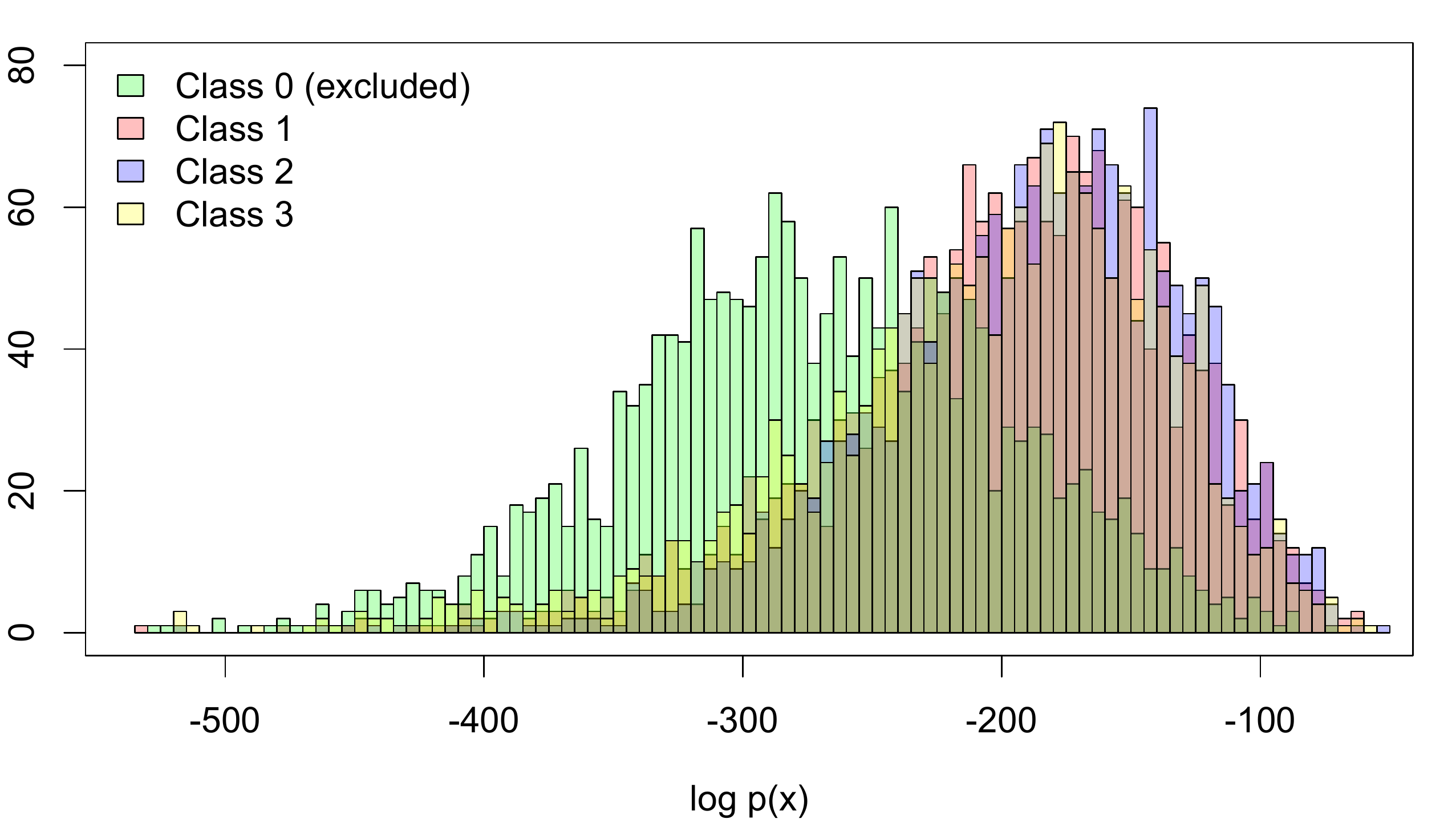}
    \includegraphics[scale=0.23]{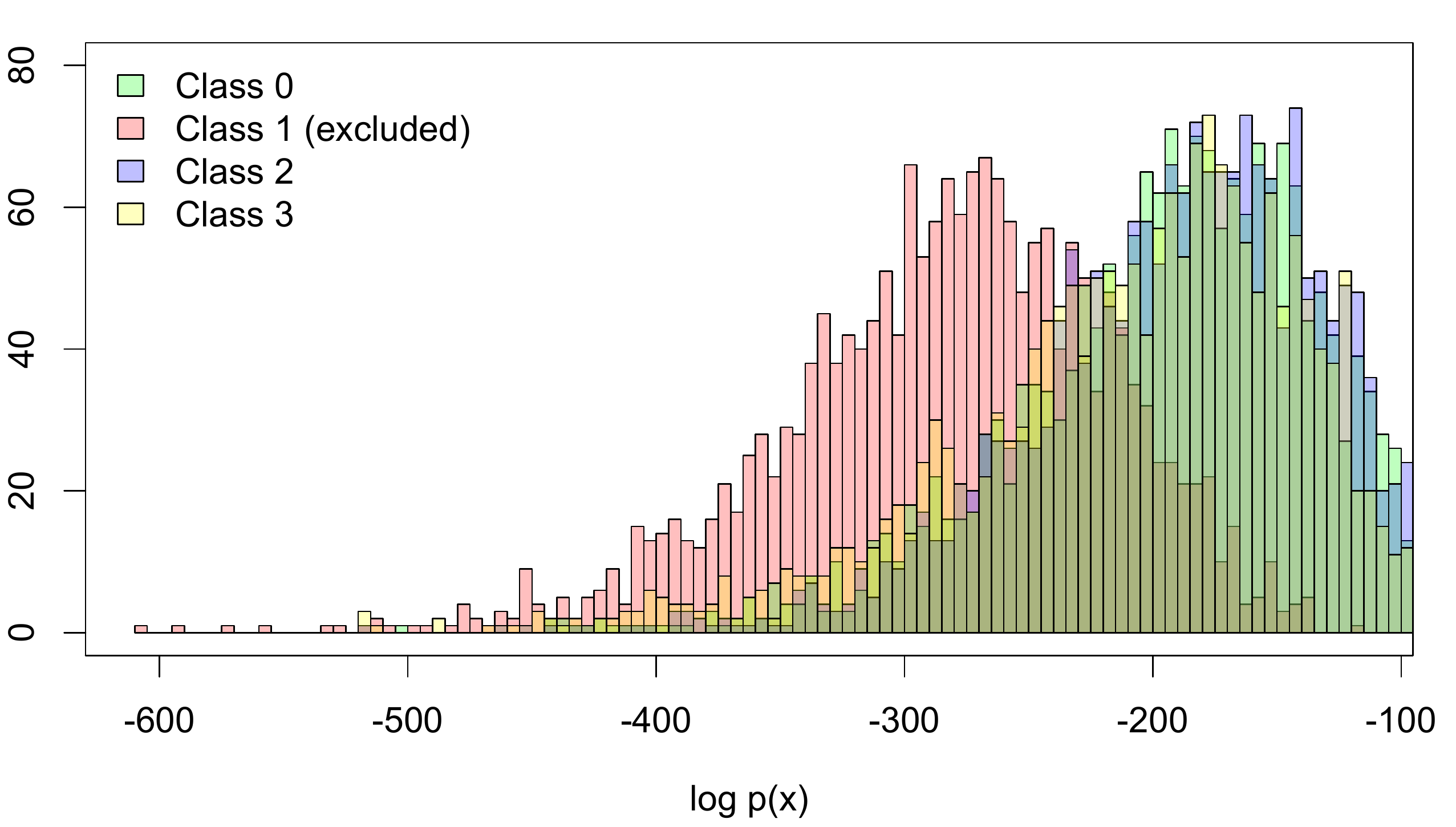}
    \vspace{-0.3cm}
    \caption{Log likelihood of test data $p(\struct{x})$ from the generative LSTM model on the AG News dataset when training data only includes three classes. In the top plot, we exclude training examples from class 0, whereas in the bottom plot we exclude training examples class 1. See text for details.}
    \label{fig:pxplot}
\vspace{-0.5cm}
\end{figure}

\vspace{-0.2cm}
\paragraph{Data likelihood.}
In generative models, we can easily compute the probability of 
a document by marginalizing over classes: 
$p(\struct{x}) = \sum_{y \in \mathcal{Y}} p(\struct{x} \mid y) p(y)$.
Since there is no explicit model of $p(\struct{x})$ in discriminative models, obtaining it would require a separate language model training.
We explore whether we can use $p(\struct{x})$ as an 
indicator of the presence of a new (unknown) class.
We use the AG News corpus and train the generative LSTM model
on examples from only 3 labels (recall that there are 4 labels
in this corpus). We compute $p(\struct{x})$
for all documents in the test set and show the results
in Figure~\ref{fig:pxplot}. 
We can see that examples from the class where
there is no training data (i.e., class 0 and class 1 in the
top and bottom figures, respectively) tend to have lower marginal likelihoods
than examples from classes observed in the training data.
In practice, we can use this observation to see
whether there is data distribution shifts 
and we need to update parameters of our models.

\section{Conclusion}
\vspace{-0.2cm}
We have compared discriminative and generative LSTM-based text classification models in terms of sample complexity and asymptotic error rates.
We showed that generative models are better than their
discriminative counterparts in small-data regime, empirically extending the (theoretical) results of
\citet{ngandjordan} from linear to nonlinear models. Formal characterization of the generalization behavior of complex neural networks is difficult, with findings from convex problems failing to account for empirical facts about generalization~\cite{generalization}. As such, this result is remarkable for being one domain in which generalization behavior of simpler models transfers to more complex models.

We also investigated their properties in the continual and zero-shot settings. Our collection of results showed that 
generative models are more suitable in these settings
and they were able to obtain comparable performance
to generative models trained on the full datasets in the
standard setting.

\bibliography{example_paper}
\bibliographystyle{icml2017}

\end{document}